\documentclass{article}

\usepackage[final, nonatbib]{nips_2016}

\usepackage[utf8]{inputenc} 
\usepackage[T1]{fontenc}    
\usepackage{hyperref}       
\usepackage{url}            
\usepackage{booktabs}       
\usepackage{amsfonts}       
\usepackage{nicefrac}       
\usepackage{microtype}      
\usepackage{amsmath}
\usepackage{graphicx}
\usepackage{subcaption}
\usepackage[backend=bibtex]{biblatex}
\usepackage{cleveref}
\usepackage[ruled]{algorithm2e}
\newcommand{\norm}[1]{\left\lVert#1\right\rVert}
\graphicspath{{./world_imgs/}{./grbm_vis/}{./vae_vis/}}

\title{Exploration for Multi-task Reinforcement Learning with Deep Generative Models}

\author{
  Sai Praveen B\\
  Department of Computer Science \& Engineering\\
  Indian Institute of Technology Madras\\
  \texttt{bpraveen@cse.iitm.ac.in} \\
  \And
  JS Suhas\\
  Department of Computer Science \& Engineering\\
  Indian Institute of Technology Madras\\
  \texttt{sauce@cse.iitm.ac.in} \\
  \And
  Balaraman Ravindran\\
  Department of Computer Science \& Engineering\\
  Indian Institute of Technology Madras\\
  \texttt{ravi@cse.iitm.ac.in} \\
}
\begin{document}
\maketitle

\begin{abstract}
Exploration in multi-task reinforcement learning is critical in training agents to deduce the underlying MDP. Many of the existing exploration frameworks such as $E^3$, $R_{max}$, Thompson sampling assume a single stationary MDP and are not suitable for system identification in the multi-task setting. We present a novel method to facilitate exploration in multi-task reinforcement learning using deep generative models. We supplement our method with a low dimensional energy model to learn the underlying MDP distribution and provide a resilient and adaptive exploration signal to the agent. We evaluate our method on a new set of environments and provide intuitive interpretation of our results.
\end{abstract}

\section{Introduction}
Learning to solve multiple tasks simultaneously is the Multi-task reinforcement learning(MTRL) problem. MTRL can be solved by either planning after deducing the current MDP or ignoring MDP deduction and learning a policy over all the MDPs combined. For example, say, a marker in the environment determines the obstacle structure and the goal locations. Any agent will have to visit the marker to learn the environment structure, and hence, deducing the task to be solved becomes an important part of the agents policy. \\
MTRL is different from Transfer learning, which aims to generalize (or \emph{transfer}) knowledge from a set of tasks to solve a new, but similar, task. MTRL also involves learning a common representation for a set of tasks, but makes no attempt to generalize to new tasks.\\
Conventional reinforcement learning algorithms like Q-learning, SARSA etc fail to identify this decision making sub-problem. They also learn sub-optimal policies because of the non-stationary reward structure since the goal location varies from episode to episode. To address these issues, current research in MTRL is driven towards using a model over all possible MDPs to deduce the current MDP or using some form of memory to store past observations and then reason based on this history. These methods will be able to deduce the MDP if they happen to see the markers, but make no effort to actively search and deduce the MDP. \\
Our methods incentivize the agent to actively seek markers and deduce the MDP by providing a smooth, adaptive exploration signal, as a \emph{pseudo}-reward, obtained from a generative model using deep neural networks. Since our method relies on computing the Jacobian of some state representation with respect to the input, we use deep neural networks to learn this state representation. To the best of our knowledge, we are the first to propose using the Jacobian as an exploration bonus. \\
Our exploration bonus is added to intrinsic reward like in Bayesian Exploration Bonus. We focus on grid-worlds with colors for markers. For clarity, we redefine a state \emph{s}, to be a single grid-location and its pixel value, \emph{$x_s$} , as the observation. Our methods can, however, generalize to arbitrarily complex distributions on the observations \emph{X}. We also assume that the agent can deduce rewards and transition probabilities from the observation.\\

\section{Related Work}
There is extensive research in the field of exploration strategies for reducing uncertainty in the MDP. $R_{max}$, $E^3$ are examples of widely used exploration strategies. Bayesian Exploration Bonus assigns a pseudo reward to states calculated using frequency of state visitation. Thomson sampling samples an MDP from the posterior distribution computed using evidence(rewards and transition probabilities) that it obtains from trajectories. We follow a similar approach to sample MDPs. Recent advances in this domain involve sampling using Dropout Neural Networks\cite{gal2015dropout}. However, these algorithms assume there exists a single stationary MDP for each episode(which we refer to as Single-Task RL). Our algorithm addresses the Multi-task RL problem where each episode uses an MDP sampled from an arbitrary distribution on MDPs. Contrary to the STRL exploration strategies, our exploration bonus is designed to mark states potentially useful for the agent to improve its certainty about the current MDP. \\
Recent advances in MTRL and Transfer Learning algorithms like Value Iteration Networks\cite{tamar2016value} and Actor-Mimic Networks\cite{parisotto2015actor} that attempt to identify common structure among tasks and generalize learning to new tasks with similar structure. In the context of MTRL and Transfer Learning on environments that give \emph{image}-like observations, Value Iteration Networks employ Recurrent Neural Networks for value iteration and learn kernel functions $f_R(x)$ and $f_P(x)$ to estimate the reward and transition probabilities for a state from its immediate surroundings. This has the effect of easily generalizing to new tasks which share the same MDP structure($R$ and $P$ for a state can be determined using locality assumptions). Our work does not attempt to learn common structure across MDPs for the purpose of transfer learning. Instead, we attempt to learn the input MDP distribution to deduce the current MDP given the observations. \\
\cite{oh2016control} proposes a novel method using Deep Recurrent Memory Networks to learn policies on Minecraft multi-task environments.
They used a fixed memory of past observations. At each step, a context vector is generated and the memory network is queried for relevant information. This model successfully learns policies on I-shaped environments where the color of a marker cell determines the goal location. In their experiments with the I-world and Pattern-Recognition world, the identifier states are very close to the agents starting position. \\
Another class of MTRL algorithms focuses on deducing the current MDP using Bayesian Reasoning. Multi-class models proposed by \cite{lazaric2010bayesian} and \cite{wilson2007multi},  attempt to assign class labels to the current MDP given a sequence of observations made from it. \cite{wilson2007multi} use a Hierarchical Bayesian Model(HBM) to learn a conditional distribution over class labels given the observations. The agent samples an MDP from the posterior distribution in a manner similar to Thomson sampling, and then chooses the action. We follow the same procedure for action selection, but incorporate exploration bonuses into it as well. \\ \cite{lazaric2010bayesian} proposes Multi-class Multi-task Learning(MCMTL), a non-parametric Hierarchical Bayesian Model to learn inherent structure in the value functions of each class. MCMTL clusters MDPs into classes and learns a posterior distribution over the MDPs given observed evidence. This is similar to our work, but it does not explicitly incentivize the agent to visit marker states. \\

Our contributions are two fold. First, we propose a deep generative model to allow sampling from posterior distribution. Second, we propose a novel exploration bonus using the models posterior distribution.

\section{Background}
\subsection{Variational Auto Encoders}
Variational Auto Encoders(VAE)\cite{kingma2013auto} attempt to learn the distribution that generated the data $\mathbf{X}$, $p(\mathbf{X})$. VAEs, like standard autoencoders have an encoder, $\mathbf{z}=f_e(\mathbf{x})$, and a decoder $\mathbf{y}=f_d(\mathbf{z})$ component. Generative models that attempt to estimate $p(\mathbf{X})$ use a likelihood objective function, $p_\theta(\mathbf{X})$ or $\log p_\theta(\mathbf{X})$. More formally, the objective function can be written as
\begin{align*}
    p(\mathbf{x}) &= \int_\mathbf{Z} p(\mathbf{x}|\mathbf{z};\theta)p(\mathbf{z})\approx \sum_{z\sim\mathbf{Z}} p(\mathbf{x}|\mathbf{z};\theta)p(\mathbf{z}) \\
    \hat{p}(\mathbf{x}) &\approx \sum_{z\in\mathbf{D}}p(\mathbf{x}|\mathbf{z};\theta)p(\mathbf{z})\,\, \text{where}\, D=\{\mathbf{z_1},\mathbf{z_2}\ldots,\mathbf{z_m}\} \,\text{and}\, \mathbf{z}_k\sim\mathbf{Z} \,\forall k
\end{align*}
where $p(\mathbf{x}|\mathbf{z};\theta)$ is defined to be $\mathcal{N}(f(\mathbf{z};\theta),\sigma^2\mathbb{I})$.  \\
Gradient-motivated learning requires approximation of the integral with samples. In high-dimensional $\mathbf{z}$-space, this could lead to large  estimation errors as $p(\mathbf{x}|\mathbf{z})$ is likely to be concentrated around a few select $\mathbf{z}$s and it would take an infeasible number of samples to get proper estimate. VAEs circumvent this problem by introducing a new distribution $\mathcal{Z} \sim \mathcal{N}(\mu_\phi(\mathbf{z}),\sigma_\phi(\mathbf{z}))$ to sample $D$ from. To reduce parameters, we use $\sigma_\phi(\mathbf{z}) = c\cdot\mathbb{I}$. These two functions are approximated with a deep network and form the \textit{encoder} component of the VAE. $p_\theta(\mathbf{X})$ is represented using the sampling function $f(z;\theta)$ where $z\sim\mathcal{N}(0,1)$ and $f(z)$ forms the \textit{decoder} component of the VAE.
After some mathematical sleight of hand to account for $\mathcal{Z}$ in the learning equations(\cite{walker2016uncertain} provides an intuitive understanding of these equations), we obtain the following formulation of the loss function
$$E=\mathcal{D_{KL}}(\mathcal{N}(\mu(\mathbf{X}),\sigma(\mathbf{X})),\mathcal{N}(0,1)) + \sum_{k=1}^N \norm{\frac{(\mathbf{y}_i-\mathbf{x}_i)}{\sigma}}^2$$
where the KL-divergence term exists to adjust for importance sampling $\mathbf{z}$ from $\mathcal{Z}$ instead of $\mathcal{N}(0,1)$.

\subsection{Gaussian-Binary Restricted Boltzmann Machines}
RBMs have been used widely to learn energy models over an input distribution $p(\mathbf{X})$. RBM is an undirected, complete bipartite, probabilistic graphical model with $N_h$ hidden units, $H$,  and $N_v$ visible units, $V$. In Gaussian-Binary RBMs, hidden units are binary units(Bernoulli distribution) capable of representing a total of $2^{N_h}$ combinations, while the visible units use the Gaussian distribution. The network is parametrized by edge weights matrix $\mathbf{W}$ between each node of $V$ and $H$, and bias vectors $\mathbf{a}$ and $\mathbf{b}$ for $V$ and $H$ respectively. Given a visible state $\mathbf{v}$, \\ the hidden state, $\mathbf{h}$, is obtained by sampling the posterior given by
$$p(\mathbf{h}|\mathbf{v}) = \frac{1}{ 1+\exp^(\mathbf{W}^T\mathbf{v} + \mathbf{b}) }$$ 
Given a hidden state $\mathbf{h}$, visible state $\mathbf{v}$ is obtained by sampling the posterior given by
$$p(\mathbf{v}|\mathbf{h}) = \mathcal{N}(\mathbf{W}^T\mathbf{h} + \mathbf{a}, \mathbf{\Sigma})$$
Since RBMs model conditional distributions, conditional distributions($p(\mathbf{v}|\mathbf{h})$ and $p(\mathbf{h}|\mathbf{v})$) have a closed form while marginal and joint distributions($p(\mathbf{v})$, $p(\mathbf{h})$ and $p(\mathbf{h}, \mathbf{v})$) are impossible to compute without explicit summation over all combinations. 

Parameters are learnt using contrastive divergence(\cite{hinton2010practical}). Learning $\mathbf{\Sigma}$, however, proved to be unstable(\cite{hinton2010practical}) and hence, we treat $\sigma$ as a hyperparameter and use $\mathbf{\Sigma}=\sigma*\mathbb{I}_{N_v}$.

\section{Deep Generative Model}
\subsection{Encoding}
Let us consider the nature of our inputs. We have assumed that the agents observed surroundings are embedded on a map as an image $X$. A mask $M$ is a binary image, of the same dimensions, with $m_i = \mathbf{1}$ if its corresponding state has been observed by the agent. We denote the $i^{th}$ pixel and mask be denoted by $x_i$ and $m_i$ respectively.

\begin{figure}[h]
    \centering
    \begin{subfigure}{0.5\textwidth}
    \centering
        \includegraphics[width=\linewidth]{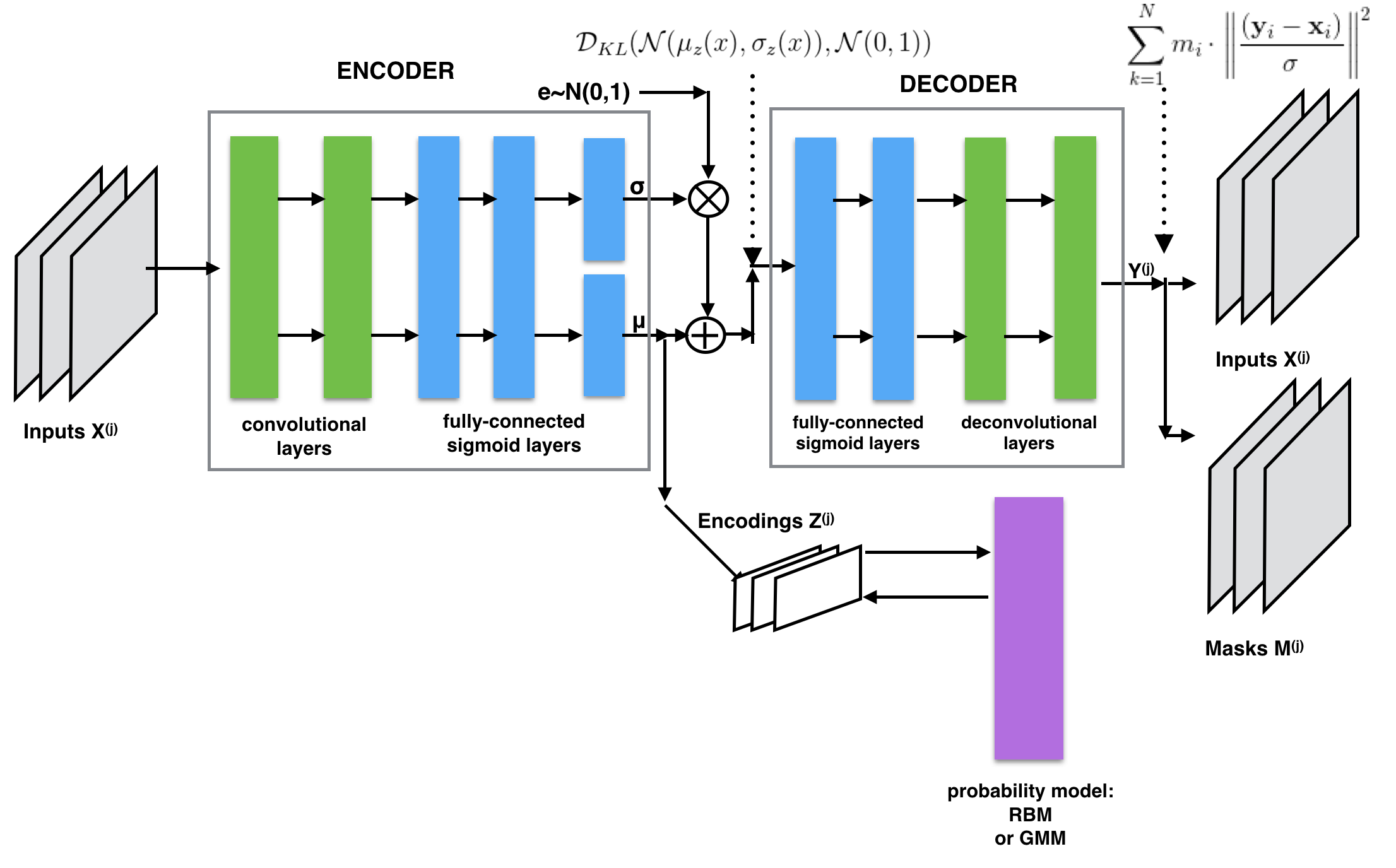}
        \caption{Train Model}
    \end{subfigure}%
    \begin{subfigure}{0.5\textwidth}
    \centering
        \includegraphics[width=\linewidth]{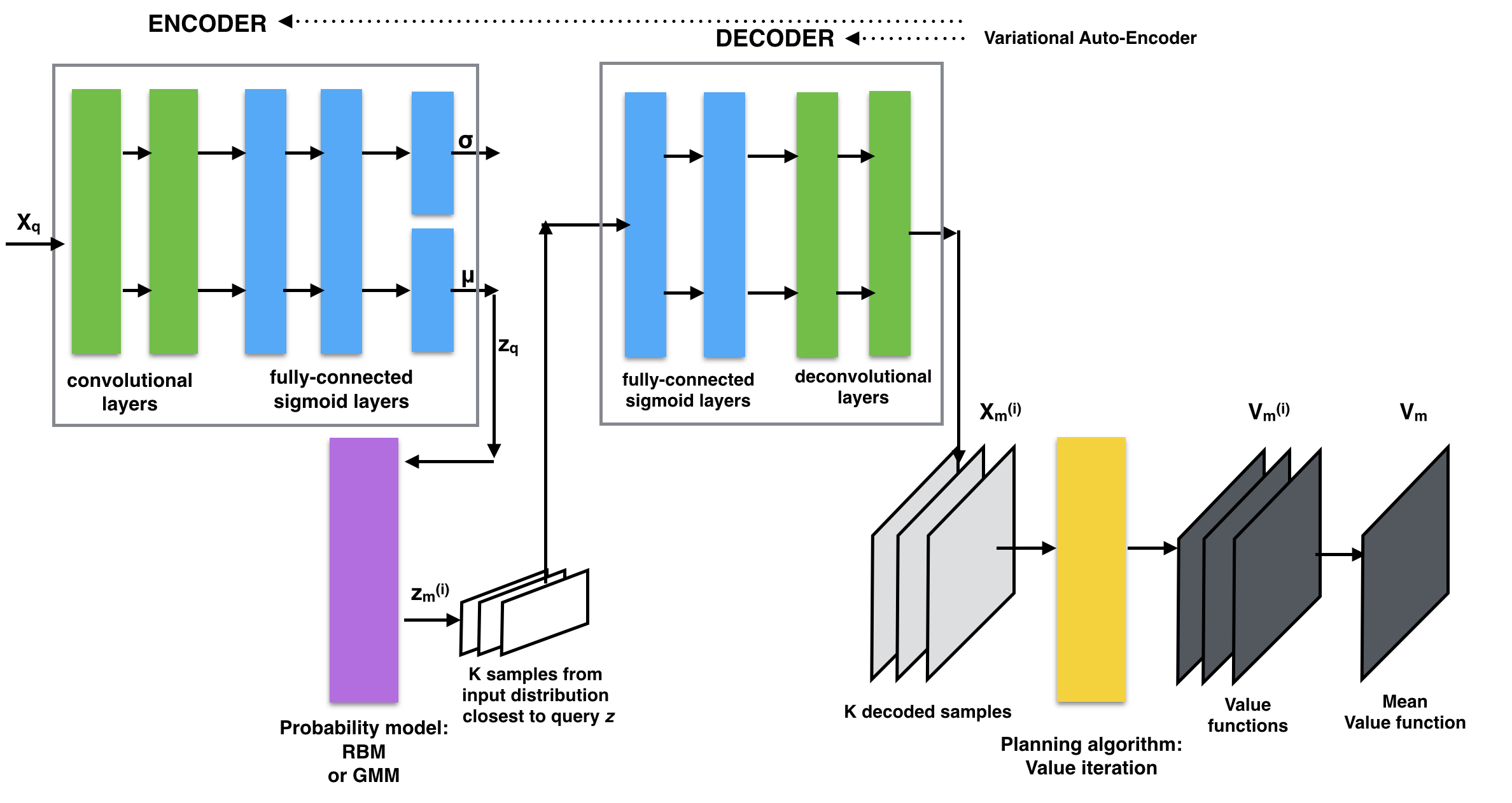}
        \caption{Query Model}
    \end{subfigure}%
    \caption{\textbf{Deep Generative Model} - Train model requires mask inputs to account for missing observations. Query model involves value iteration to determine best action over sampled MDPs.}
    \label{deep-gen-model}
\end{figure}

In most episodes, the agent will not visit the entire grid-world, hence $m_i = 0$ for some $x_i \in X$. Since there can be several views of the same ground-truth MDP, we need to be able to reconstruct the ground-truth MDP from multiple observations of the MDP over several episodes. For Single-Task RL, this can be done in a tabular fashion. In MTRL, however, we have potentially infinite  possible MDPs and it becomes hard to build association between different views of the same MDP. \\
We use deep convolutional VAEs to infer the association between different views of the same MDP and use it with a low dimensional energy model to sample MDPs given observational evidence. Here, locality of the world features warrant the use of convolution layers in the VAE. Figure \ref{deep-gen-model} shows our setup to learn the associations and to infer ground truth MDP given observations. We use one setup to train the model and another to allow back sampling of MDPs. We call these the \emph{train} and \emph{query} models. \\
Our method can be scaled to large state spaces because of the VAE.

Given this setup, for the learning phase, we modify the VAE loss function to account for unobserved states, $x_i \in X$ with $m_i=0$. The new loss function is given by
$$E=\mathcal{D_{KL}}(\mathcal{N}(\mu(\mathbf{X}),\sigma(\mathbf{X})),\mathcal{N}(0,1)) + \sum_{k=1}^N m_i \cdot\norm{\frac{(\mathbf{y}_i-\mathbf{x}_i)}{\sigma}}^2$$
Inclusion of $m_i$ in the loss function is quite intuitive and works well on the sets that we tested on, since it removes any penalty for unseen $x_i$ and allows the VAE to project its knowledge onto the unseen states.

\subsection{Sampling}
Given a partial observation, $X$, we sample for the posterior to obtain $K$ MDP samples. If $X$ doesn't have enough evidence to skew the posterior in favour of one single MDP, then the encoding produced by VAE, $\mathbf{z}$, is far from encodings of ground-truth MDPs, $\mathbf{z}_{in}$ in $\mathbf{z}$-space. We obtain an MDP that is a mixture of MDPs if we sample from this posterior. Solving this MDP could result in the agent following a policy unsuitable for any of the component MDPs in isolation. \\
One way to circumvent this problem is to train a probability distribution over the MDP embeddings, $\mathbf{z}_{in}$. For our 2-MDP environments, we use a Gaussian-Boltzmann RBM to cluster inputs with fixed-variance gaussians. We then use Algorithm \ref{sample-algo} to sample from these gaussians.
\begin{algorithm}[]
    \SetAlgoLined
    \KwResult{$K$ MDPs sampled from model posterior}
        Compute $\mathbf{z} = f_e(\mathbf{x})$\\
        Sample $\mathbf{K}$ hidden RBM states $\mathbf{h^{(i)}} \in H$, $i \in \{1\ldots K\}$ from the posterior $p(\mathbf{h}|\mathbf{z})$ \\
        Calculate MAP estimate $\mathbf{z^{(i)}} = \arg\max_\mathbf{z} p(\mathbf{z}|\mathbf{h^{(i)}})$ \\
        Decode map estimates $\mathbf{z^{(i)}}$ to get MDP samples $\mathbf{y^{(i)}}$ \\
    \caption{Sample MDPs given $\mathbf{x}$}
    \label{sample-algo}
\end{algorithm}

\subsection{Value function}
Given $K$ samples from model posterior, $p(\mathbf{y}|\mathbf{x})$, we perform action selection using an aggregate value function over the $K$ samples. We define, for each state $s$, an aggregate value function $\bar{V}(s)$ as
\begin{align*}
    \bar{V}(s) = \mathbb{E}_{\mathbf{m}\sim p(\mathbf{y}|\mathbf{x})} \big[ V_{\mathbf{m}}(s)\big] \approx \frac{\sum_{k=0}^K V_{\mathbf{m}_k}(s)}{K}
\end{align*}
where $\mathbf{m}$ is an MDP and $V_\mathbf{m}(s)$ denotes the value function for state $s$ under MDP $\mathbf{m}$. $V_\mathbf{m}(s)$ can be obtained using any standard planning algorithms and we use value iteration(with $\gamma$=0.95, 40 iterations). Action selection is done using $\epsilon$-greedy mechanism with $\epsilon=0.1$. Since recomputing value functions at each step is computationally infeasible, each selected action persists for $\tau=3$ steps. \\
We note that value functions used need not be exact, but can be approximate as they are only used for $\tau$ steps. A quicker estimate can be obtained using Monte-Carlo methods when the state-space is large.

\section{Jacobian Exploration Bonus}
To incentivize the agent to visit decisive pixels/locations, we introduce a bonus based on the change in the embedding $Z$. Intuitively, the embedding $Z$ has the highest change when the VAE detects changes that are relevant to the distribution $p(X)$ that it is modelling. The bonus can be summarised as follows:
$$ B_\alpha(s) = \alpha \cdot tanh\bigg(\epsilon + \norm{\frac{\partial \mathbf{z}}{\partial \mathbf{x_s}}}\bigg)$$
where $\mathbf{x}_s$ denotes the list of observations made at state $s$. We use a $tanh(\cdot)$ transfer function to bound activations produced by the Jacobian, thus mainitaining numerical stability.
This bonus can be used in two ways - as a pseudo reward,
$$ R_{s} = R_{s}(\mathbf{x}) + B_\alpha(s) $$ 
or to replace the actual reward.
$$ R_{s} = max( R_{s}(\mathbf{x}),B_\alpha(s) ) $$ 
where $R_{s}(\mathbf{x})$ is the actual reward deduced by the agent.
While both methods showed improvement, the latter worked better since total reward for states which already gave a high reward was not further increased.
Since $B_\alpha$ changes drastically with new observations,
$B_\alpha$ is recomputed every time the $\hat{V}_t(s)$ is to be recomputed.
$B_\alpha$ is also memory-less i.e. it doesn't carry over any information from one episode to the next.

\begin{figure}[h]
    \centering
    \includegraphics[width=0.25\linewidth]{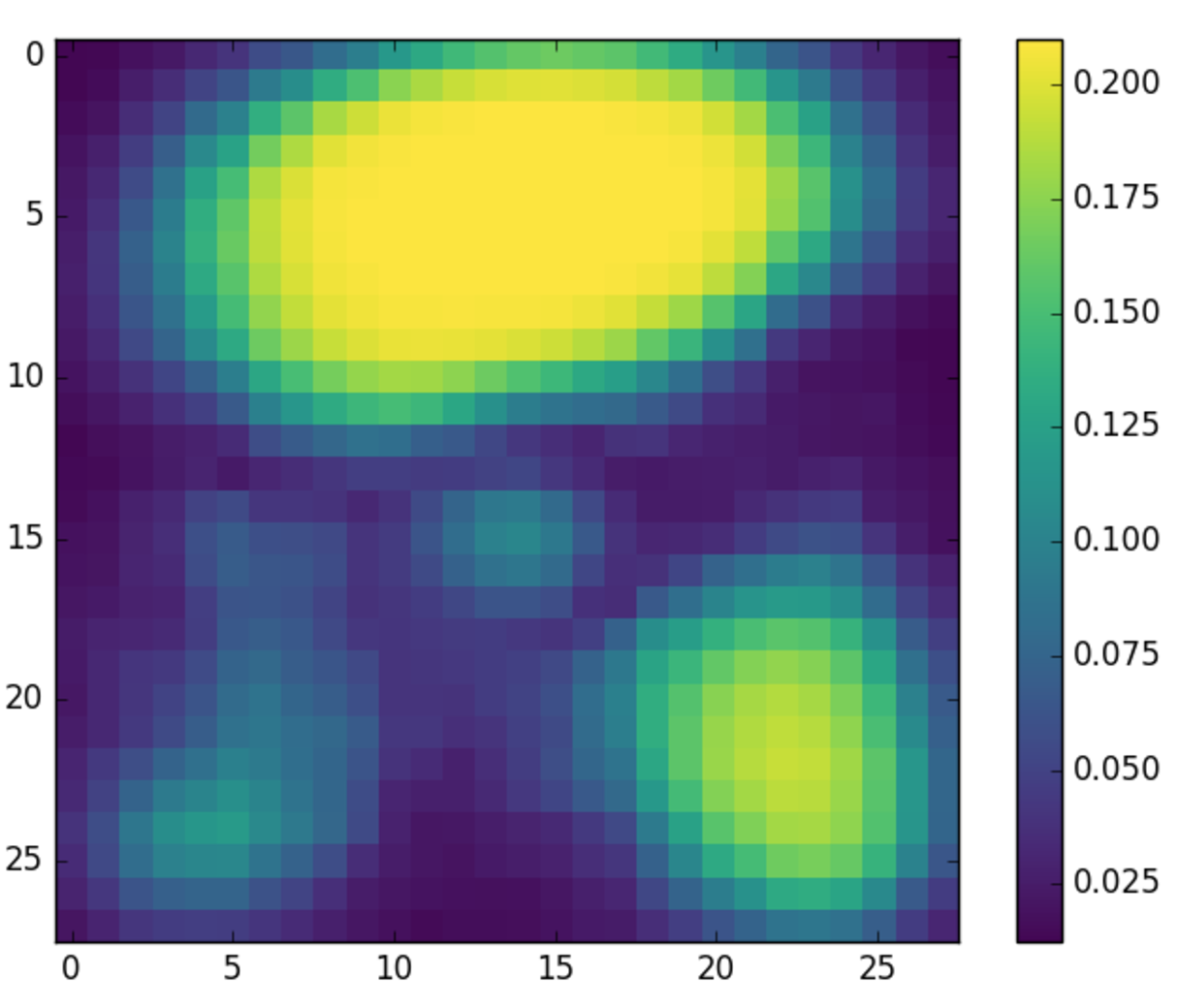}
    \quad
    \includegraphics[width=0.243\linewidth]{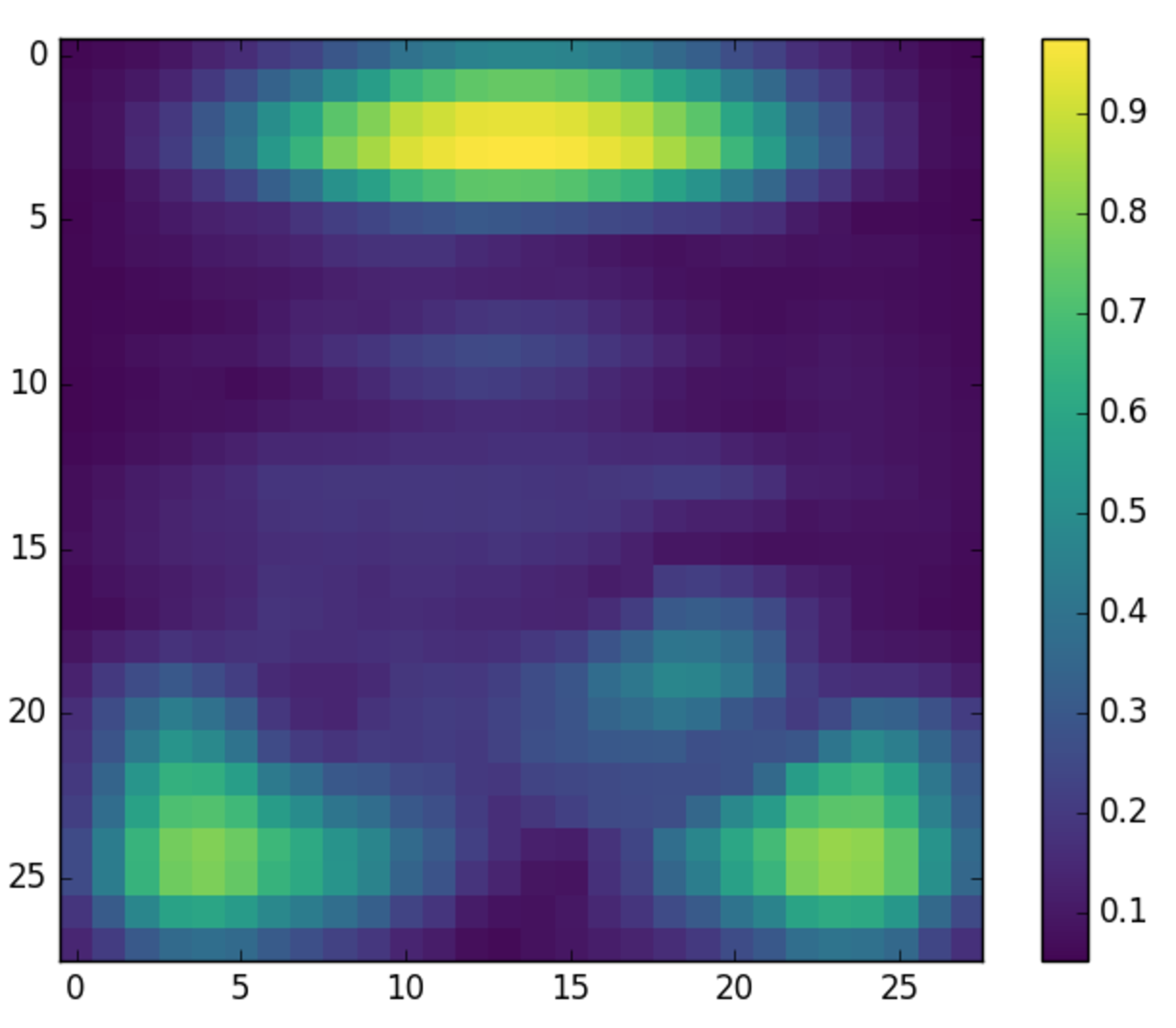}
    \caption{\textbf{Final Jacobian Bonus for BW-E and BW-H} - Locations in yellow-green are identified by the agent as being most helpful in deducing the MDP being solved.}
    \label{jacobian}
\end{figure}
\section{Experiments}
\subsection{Testbench}
We have implemented the following algorithms.
\begin{itemize}
    \item Value Iteration, referred to as STRL
    \item Multi-task RL with VAE, RBM without exploration bonus, referred to as MTRL-0
    \item Multi-task RL with VAE, RBM, and Jacobian Bonus, referred to as MTRL-$\alpha$
\end{itemize}
We have tested the above algorithms on 2 environments.
\begin{itemize}
    \item Back World (Easy) [BW-E] - Goal location alternates depending on marker location color, marker location is fixed and is in most paths from start to goal. This domain demonstrates the advantage gained using a probabilistic model over the MDPs.
    \item Back World (Hard) [BW-H] - Same setting as BW-H, but marker location is not on most paths from start to goal. This domain demonstrates the advantage provided by the Jacobian exploration bonus and our generative model.
\end{itemize}

For STRL, using only visible portions of the environment was very unstable and hence, we had to add a pseudo reward. For each unseen location, we provide a pseudo reward, $\varepsilon_n$ for $n^{th}$ step (with $\varepsilon_0 = 0.3$), that is annealed by a factor of $\kappa = 0.9$. Each episode was terminated at 200 steps if the agent hadn't reached the goal. Using this pseudo reward, the agent was forcefully terminated fewer times. 
These worlds become challenging due to partial visibility. We use a 5x5 kernel with clipped corners and the agent is always assumed to be at the center. At each step, the environment tracks the locations that the agent has seen and presents it to the agent before an action is taken. \\
For our experiments, we consider the average number of steps to goal as a measure of loss and average reward as a measure of performance.

\begin{figure}[h]
    \centering
    \begin{subfigure}{0.25\textwidth}
    \centering
        \includegraphics[width=0.95\linewidth]{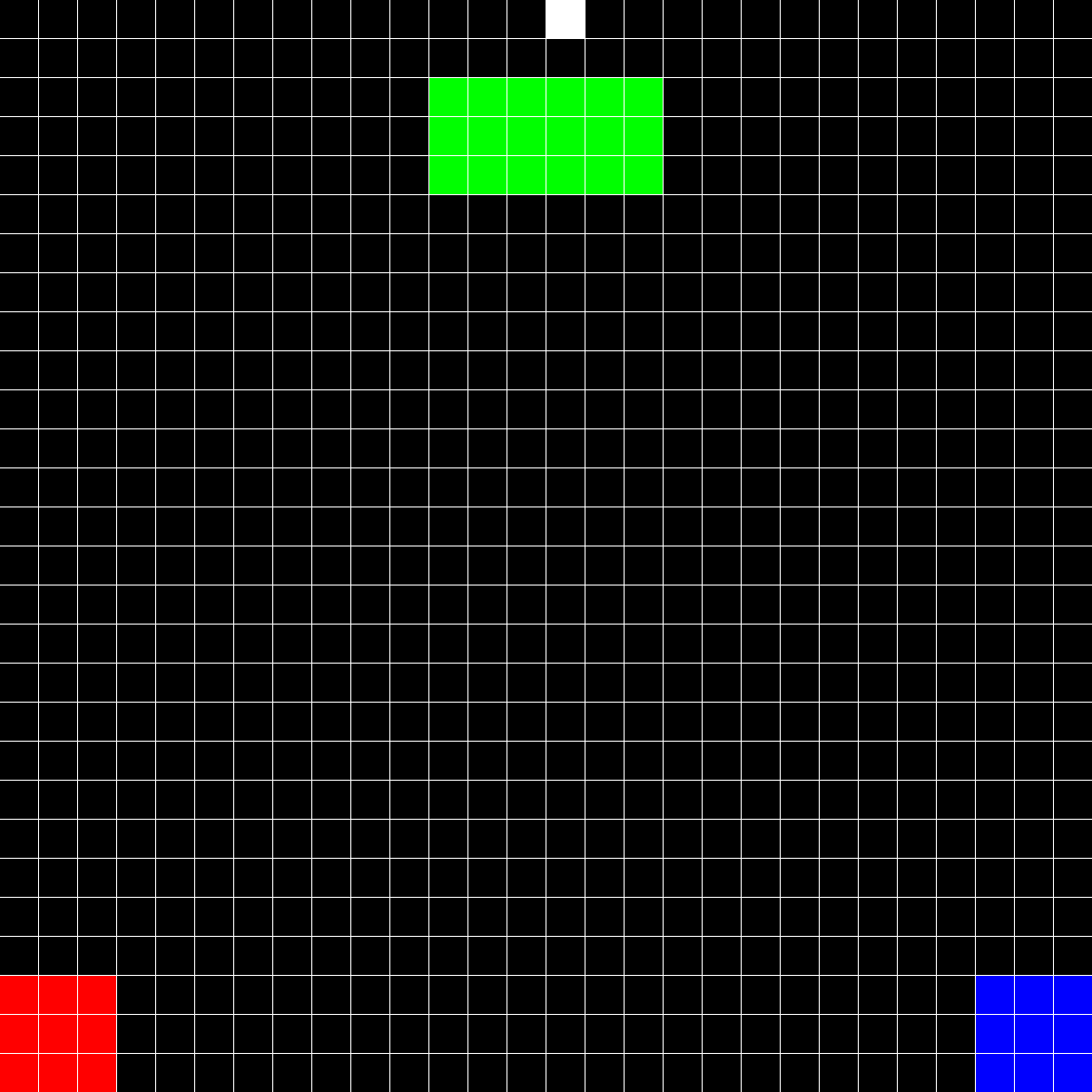}
        \caption{BW-E A}
    \end{subfigure}%
    \begin{subfigure}{0.25\textwidth}
    \centering
        \includegraphics[width=0.95\linewidth]{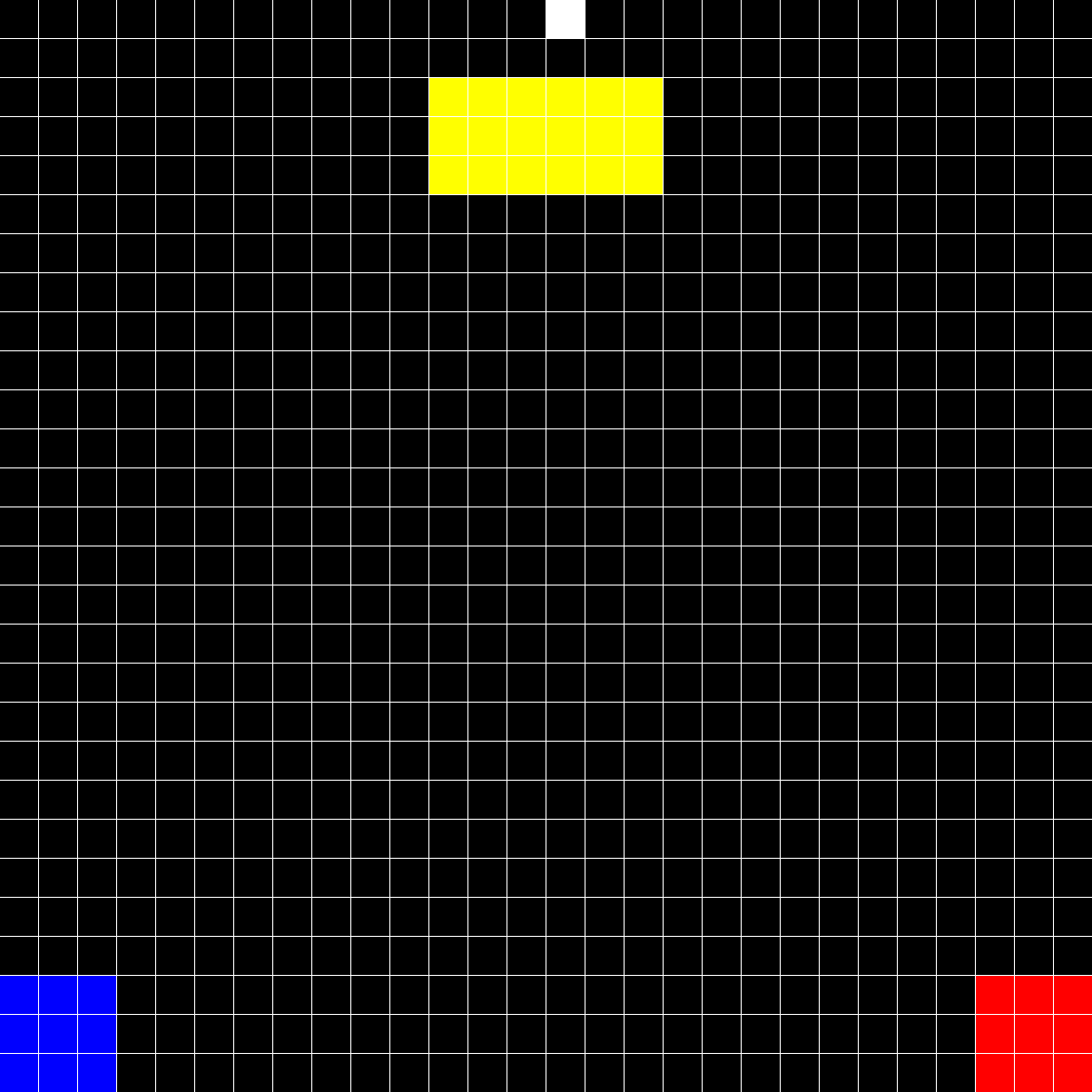}
        \caption{BW-E B}
    \end{subfigure}%
    \begin{subfigure}{0.25\textwidth}
    \centering
        \includegraphics[width=0.95\linewidth]{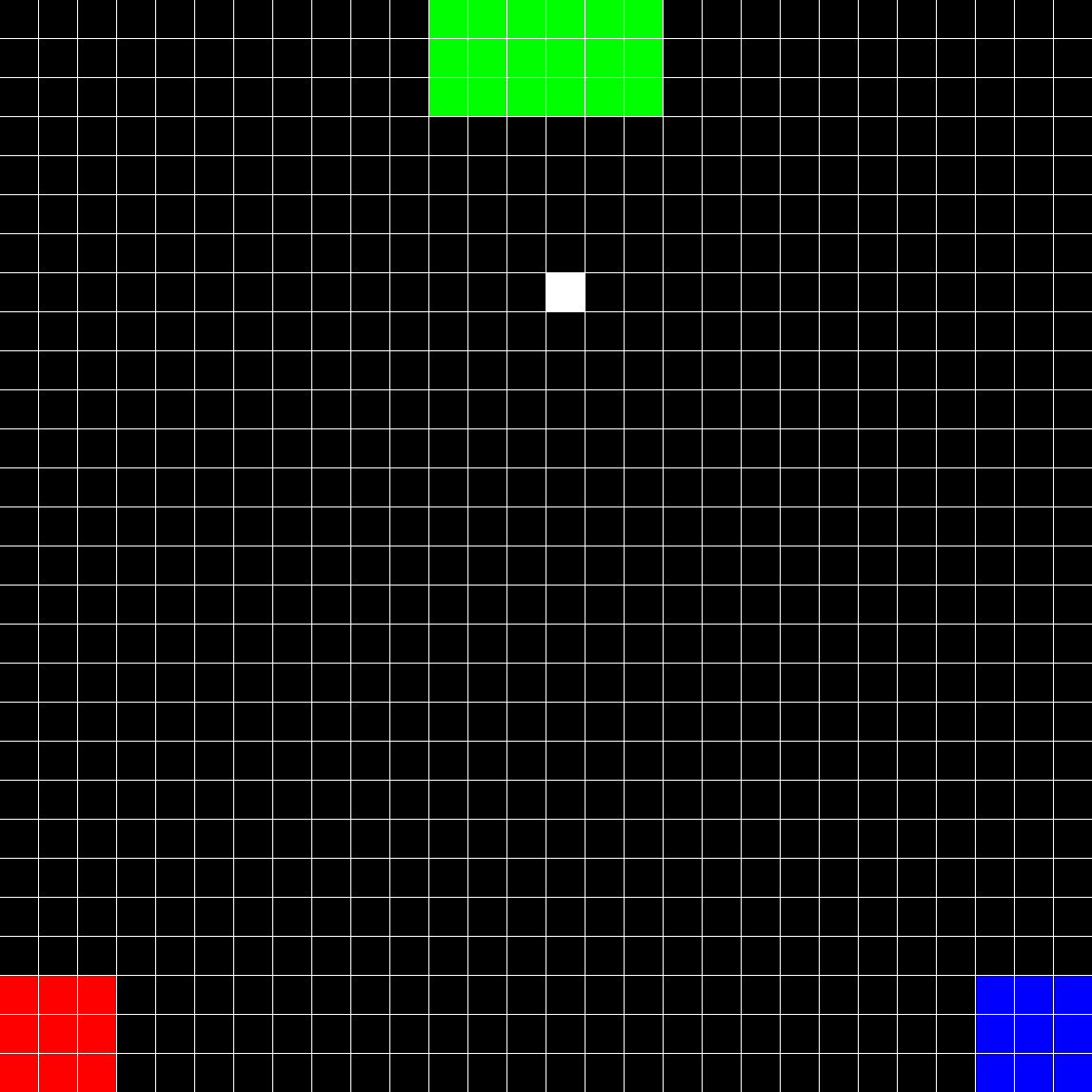}
        \caption{BW-H A}
    \end{subfigure}%
    \begin{subfigure}{0.25\textwidth}
    \centering
        \includegraphics[width=0.95\linewidth]{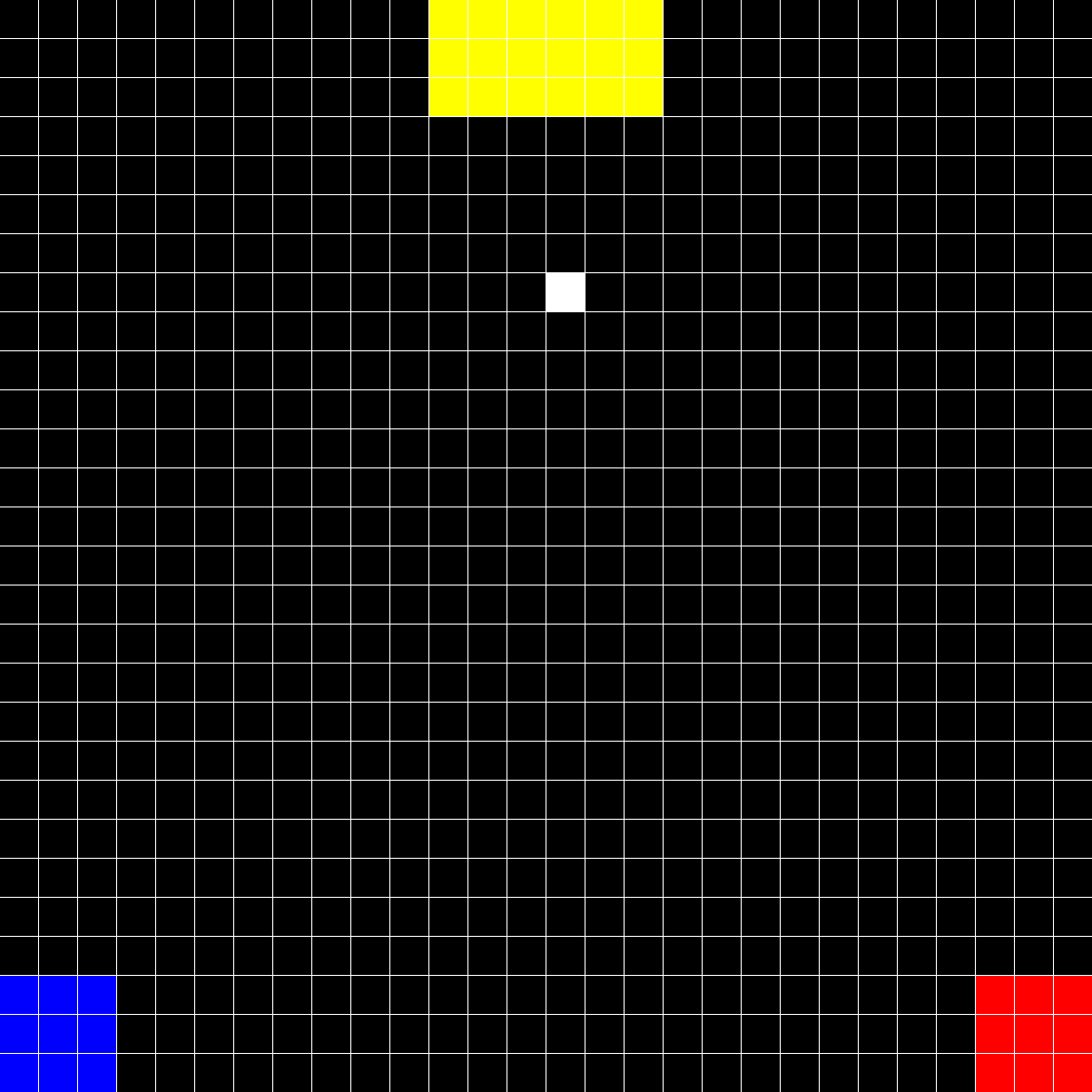}
        \caption{BW-H B}
    \end{subfigure} %
    \newline
    \begin{subfigure}{0.25\textwidth}
    \centering
        \includegraphics[width=\linewidth]{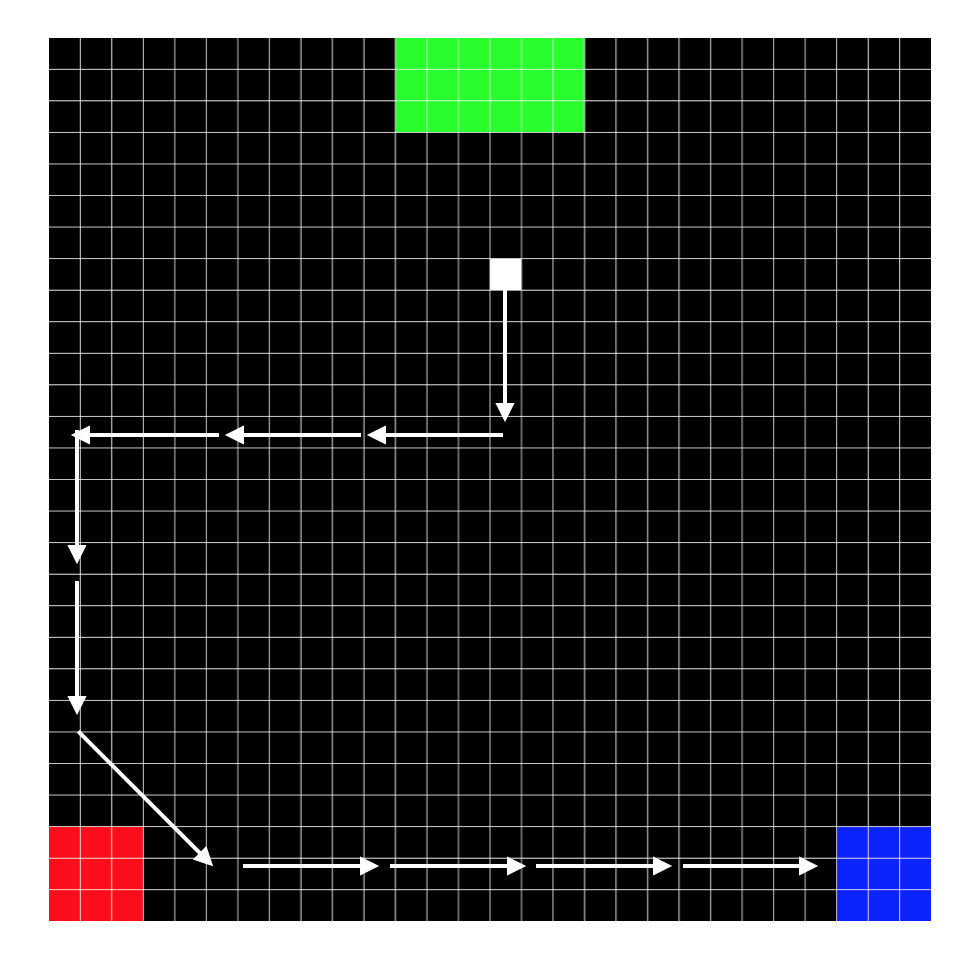}
        \caption{Suboptimal path[BW-H]}
    \end{subfigure}%
    \begin{subfigure}{0.25\textwidth}
    \centering
        \includegraphics[width=\linewidth]{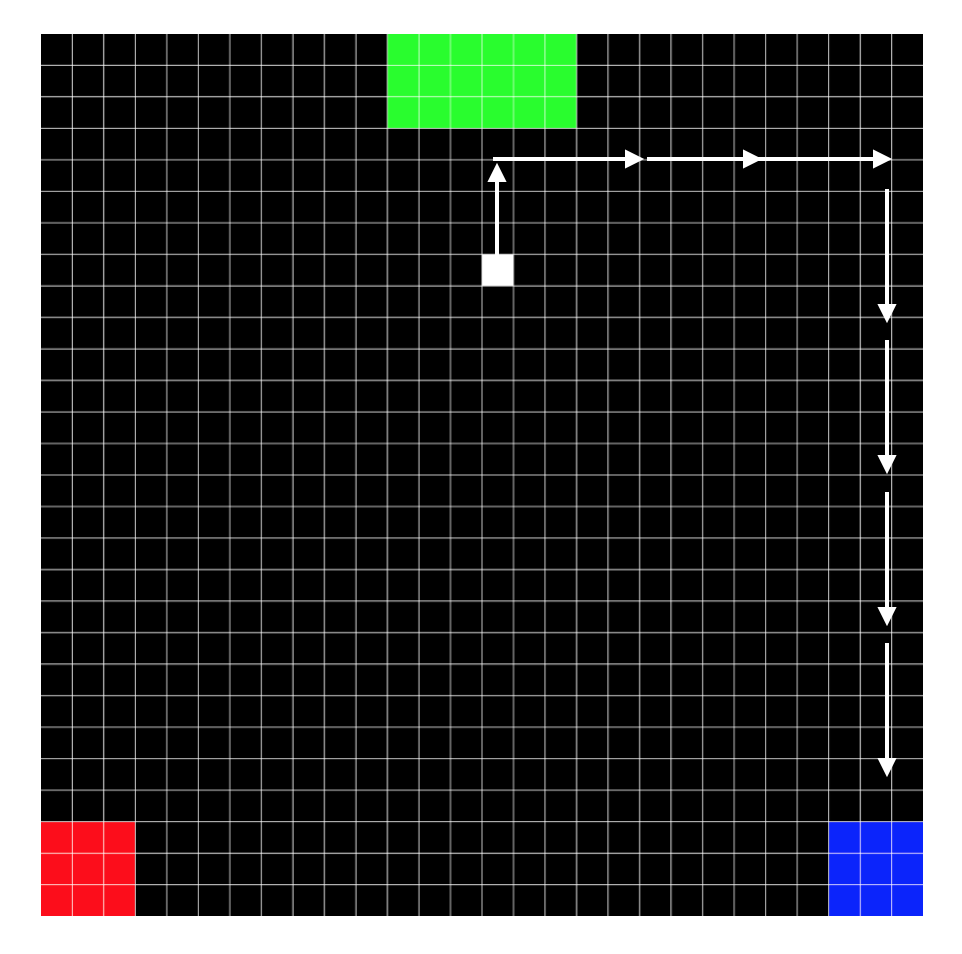}
        \caption{Optimal path[BW-H]}
    \end{subfigure}%
    \begin{subfigure}{0.25\textwidth}
    \centering
        \includegraphics[width=0.95\linewidth]{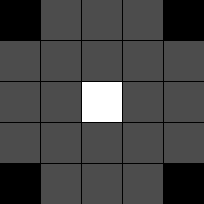}
        \caption{Visibility Kernel}
    \end{subfigure} %
    \label{envs}
    \caption{\textbf{28x28 worlds used in our experiments } - White indicates start position of agent. Green and Yellow are marker locations. Red locations are failures. Blue locations are all successes. Gray areas in kernel are visible to the agent. White cell in kernel is the agents position. Shown optimal path considers MDP deduction as a sub-problem.}
\end{figure}
 
\section{Results}
\subsection{Navigation}
Table \ref{return-table} gives average reward for each agent. Table \ref{length-table} gives average episode length. We also impose forceful termination at 200 steps if episode has not yet completed. From the results, we infer the following.
\begin{itemize}
    \item STRL using value iteration does poorly as it as no way of deducing MDPs.
    \item MTRL-0 solves both BW-E and BW-H environments and does almost as good as MTRL-$\alpha$. This improvement can be attributed to the use of our deep generative model.
    \item MTRL-$\alpha$ shows better results on BW-H. This was expected as MTRL-0 makes no attempt to visit marker locations. MTRL-$\alpha$ is motivated by the Jacobian Bonus to visit marker locations, thereby deducing the MDP.
    \item MTRL-0 performs as good as MTRL-$\alpha$ in BW-E as marker locations lie on most paths to the goal. However, since it fails to understand the significance of the marker locations and markers in BW-H are not on most paths to the goal, it results in longer episodes and lower reward.
\end{itemize}

\begin{table}[t]
\begin{minipage}{0.48\textwidth}
\centering
  \caption{Average Reward}
  \label{return-table}
  \centering
  \begin{tabular}{llll}
    \toprule
    World & STRL & MTRL-0 & MTRL-$\alpha$ \\
    \midrule
    BW-E & 0.21 & \textbf{0.99} & \textbf{0.99} \\ 
    BW-H & 0.23 & 0.92 & \textbf{0.99} \\
    \bottomrule
  \end{tabular}
  
\end{minipage}
\hfill
\begin{minipage}{0.48\textwidth}
\centering
  \caption{Average Episode Length}
  \label{length-table}
  \centering
  \begin{tabular}{llll}
    \toprule
    World & STRL & MTRL-0 & MTRL-$\alpha$ \\
    \midrule
    BW-E & 184.19 & \textbf{46.20} & 46.29 \\
    BW-H & 183.64 & 54.0 & \textbf{45.8} \\
    \bottomrule
  \end{tabular}
 \end{minipage}
\end{table}

\subsection{Visualizations}
\subsubsection{RBM Training}
\begin{figure}[h]
    \centering
    \begin{subfigure}{0.25\textwidth}
    \centering
        \includegraphics[width=\linewidth]{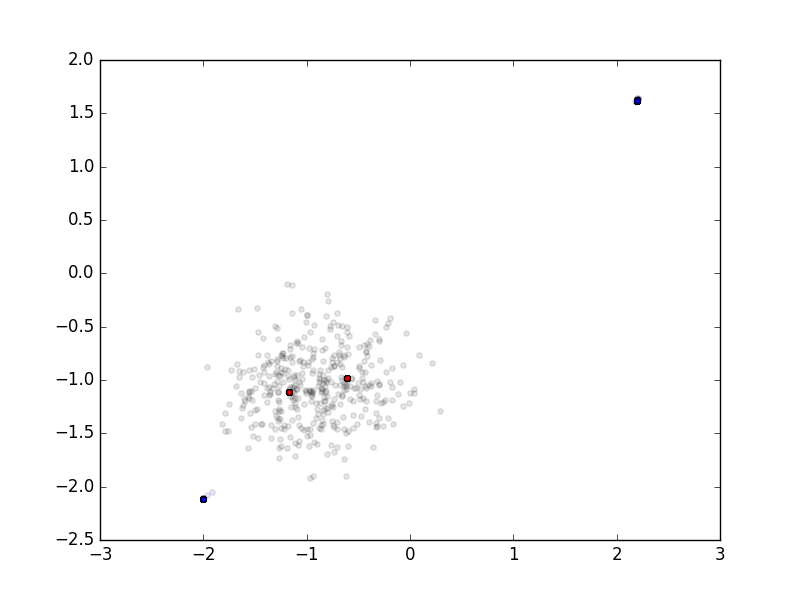}
        \caption{Epoch 0}
    \end{subfigure}%
    \begin{subfigure}{0.25\textwidth}
    \centering
        \includegraphics[width=\linewidth]{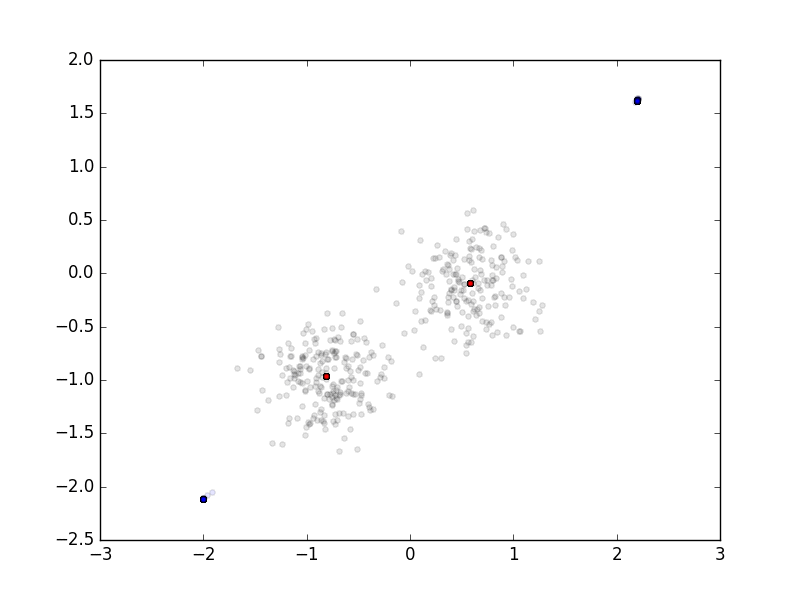}
        \caption{Epoch 3}
    \end{subfigure}%
    \begin{subfigure}{0.25\textwidth}
    \centering
        \includegraphics[width=\linewidth]{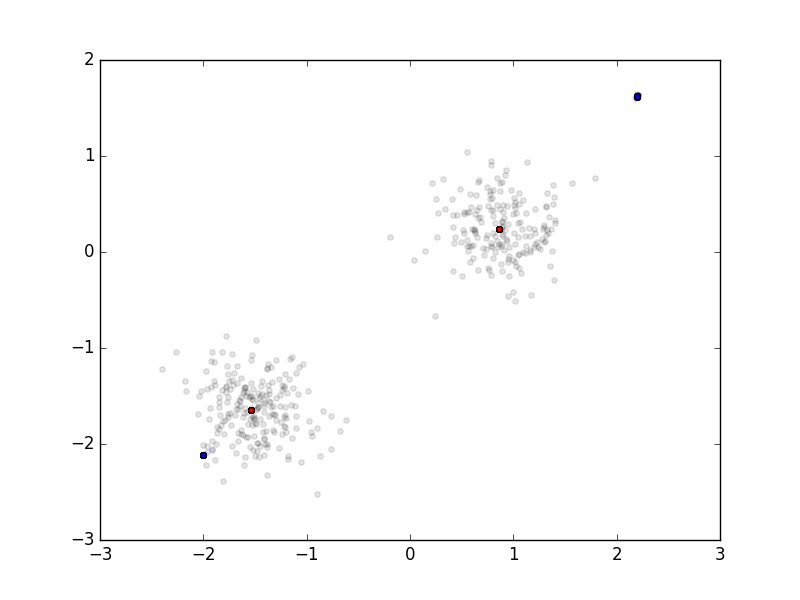}
        \caption{Epoch 6}
    \end{subfigure}%
    \begin{subfigure}{0.25\textwidth}
    \centering
        \includegraphics[width=\linewidth]{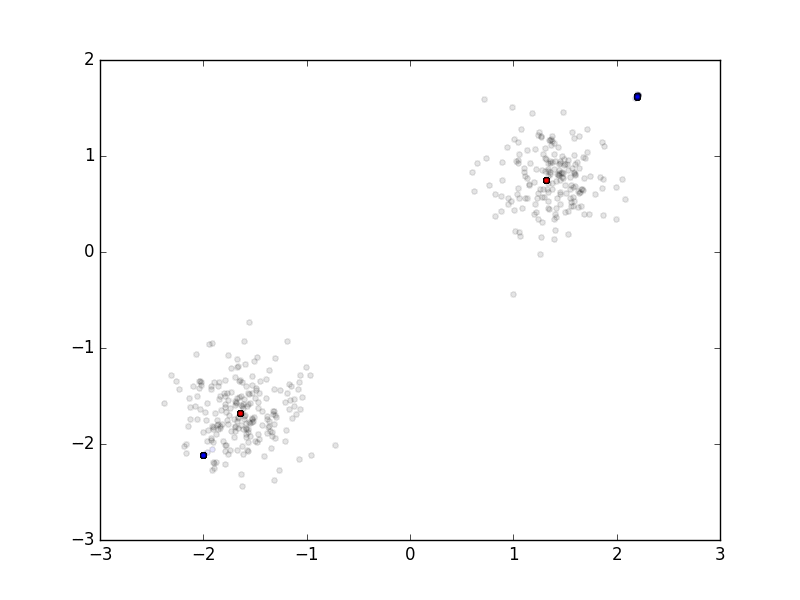}
        \caption{Epoch 9}
    \end{subfigure}%
    \newline
    \begin{subfigure}{0.25\textwidth}
    \centering
        \includegraphics[width=\linewidth]{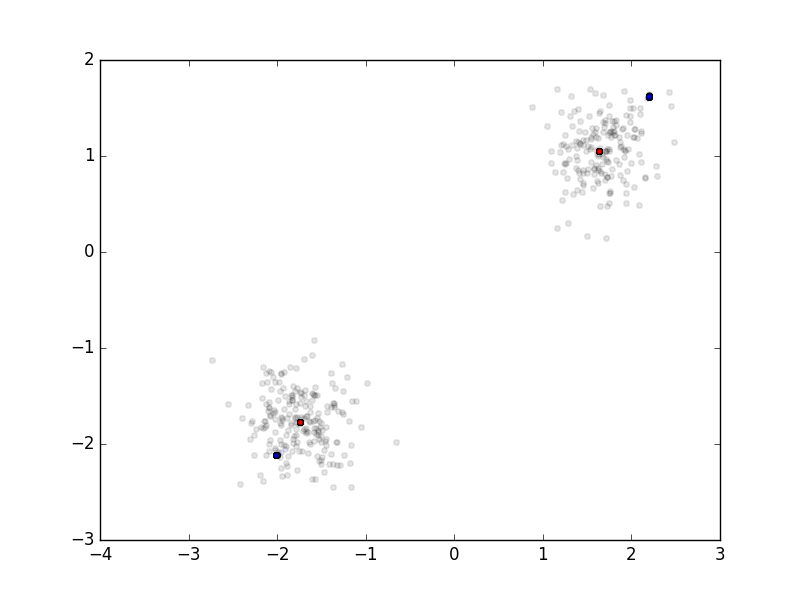}
        \caption{Epoch 12}
    \end{subfigure}%
    \begin{subfigure}{0.25\textwidth}
    \centering
        \includegraphics[width=\linewidth]{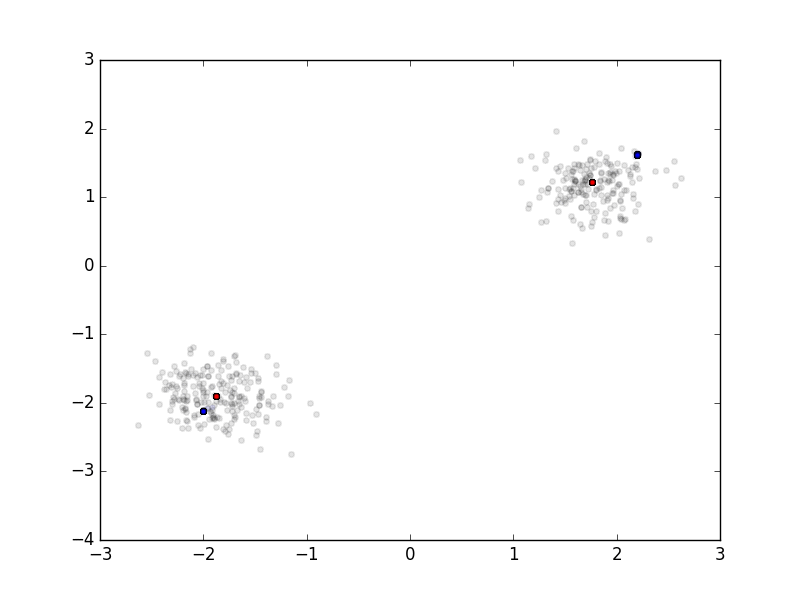}
        \caption{Epoch 15}
    \end{subfigure}%
    \begin{subfigure}{0.25\textwidth}
    \centering
        \includegraphics[width=\linewidth]{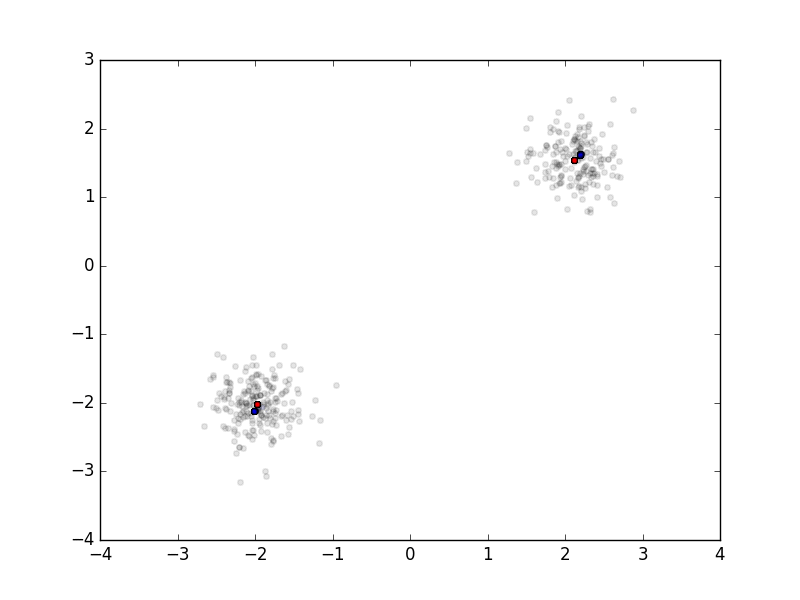}
        \caption{Epoch 26}
    \end{subfigure}%
    \begin{subfigure}{0.25\textwidth}
    \centering
        \includegraphics[width=\linewidth]{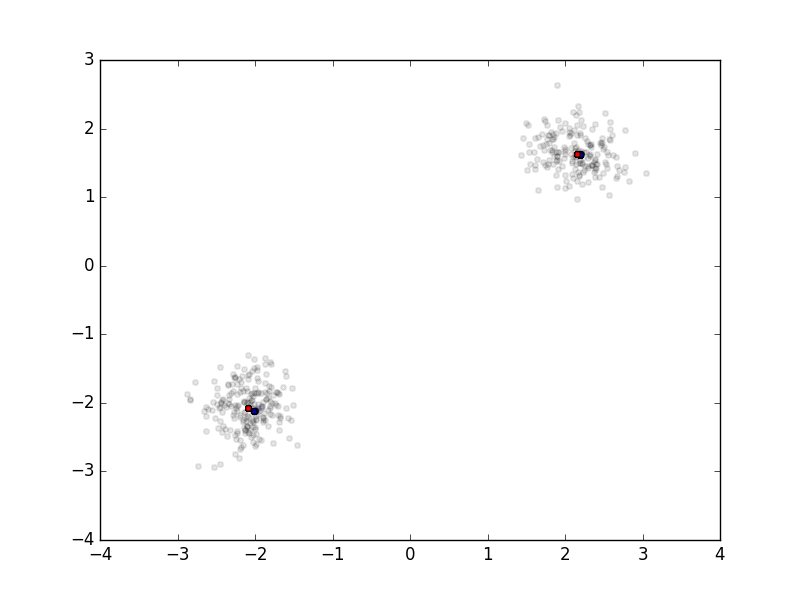}
        \caption{Epoch 79}
    \end{subfigure}%
    \caption{\textbf{Visualization of RBM training on encoded BW-E world samples} - Red points are means of the fitted gaussians. Blue points are data points in the minibatch. Black points are samples from the gaussians fitted by RBM. Spread of the black points is a measure of the variance of the fitted gaussians.}
    \label{fig:grbm-vis}
\end{figure}

 To visualize RBM training, we used the BW-E environment. We used a random agent,($\epsilon$-greedy policy with $\epsilon$=1 and restricted actions at boundaries so it doesn't bump into walls) on this environment to navigate and collect a \emph{sample}, of the seen environment, at the end of each episode. We then encoded the samples using VAE and used it to train a RBM. Figure \ref{fig:grbm-vis} shows the clusters and their means for each gaussian fit by the RBM. Since there are only two possible BW-E environments, RBM only fits two gaussians and our encoded samples are clustered around the same. Training was done in minibatches of 64 samples with 1 hidden unit in the RBM.
\\We see two distinct clusters in each snapshot. RBM iteratively refines its parameters to fit the means close to the encoded sample clusters. Due to perfect reconstruction from VAE, there is no spread of encoded samples.

\subsubsection{VAE Training}
\begin{figure}[h]
    \centering
    \begin{subfigure}{0.33\textwidth}
    \centering
        \includegraphics[width=0.95\linewidth]{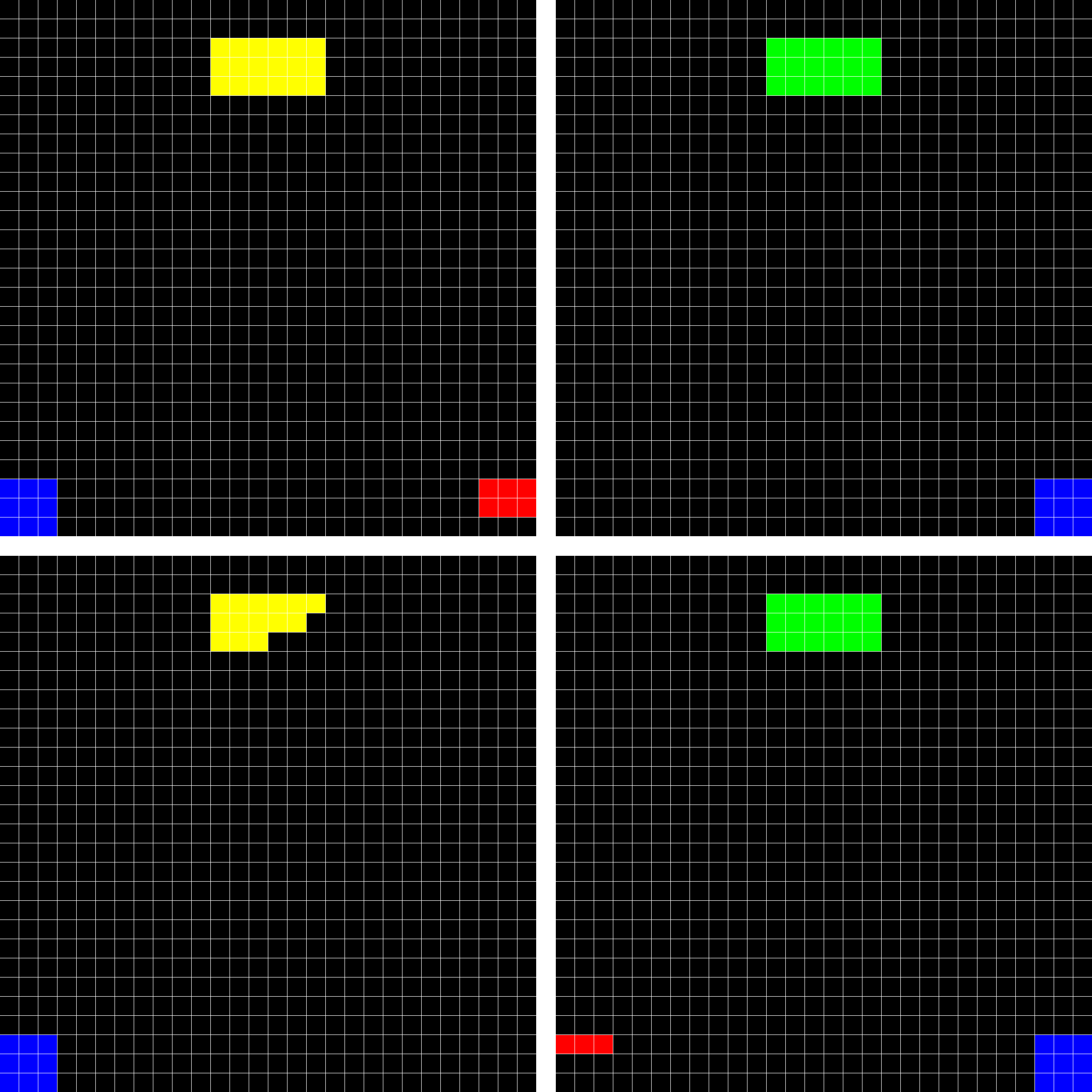}
        \caption{Input}
    \end{subfigure}%
    \begin{subfigure}{0.33\textwidth}
    \centering
        \includegraphics[width=0.95\linewidth]{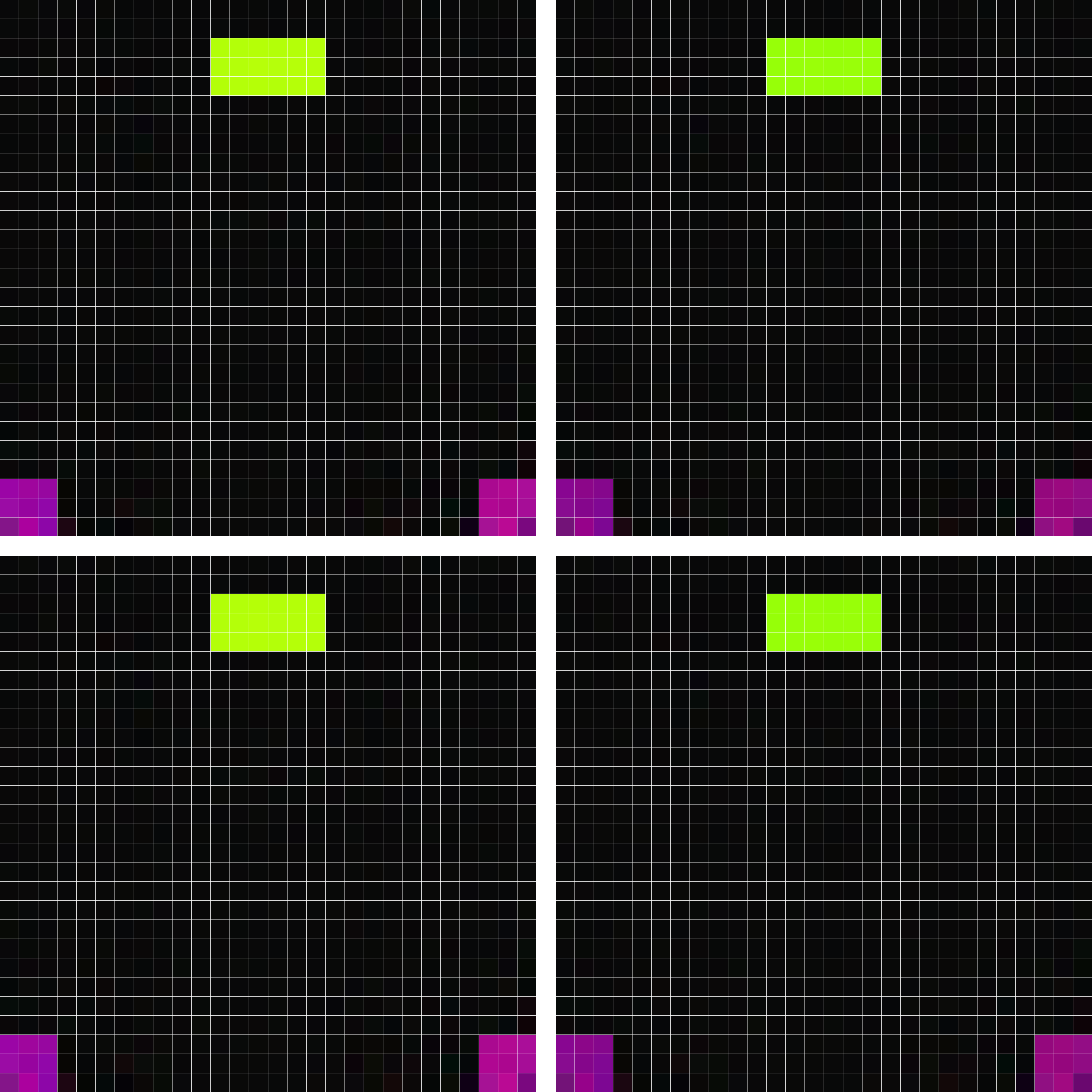}
        \caption{Epoch 30}
    \end{subfigure}%
    \begin{subfigure}{0.33\textwidth}
    \centering
        \includegraphics[width=0.95\linewidth]{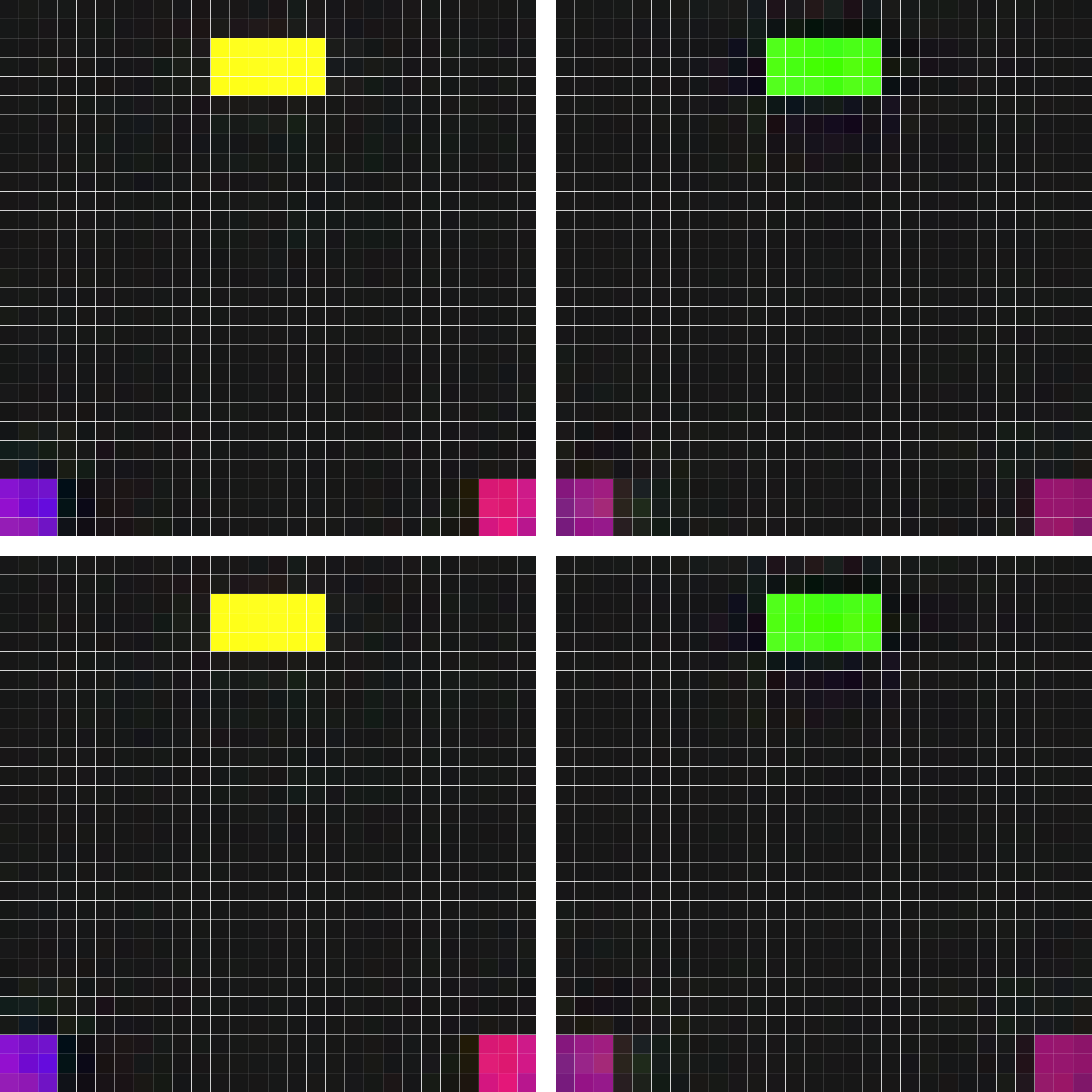}
        \caption{Epoch 60}
    \end{subfigure}%
    \newline
    \begin{subfigure}{0.33\textwidth}
    \centering
        \includegraphics[width=0.95\linewidth]{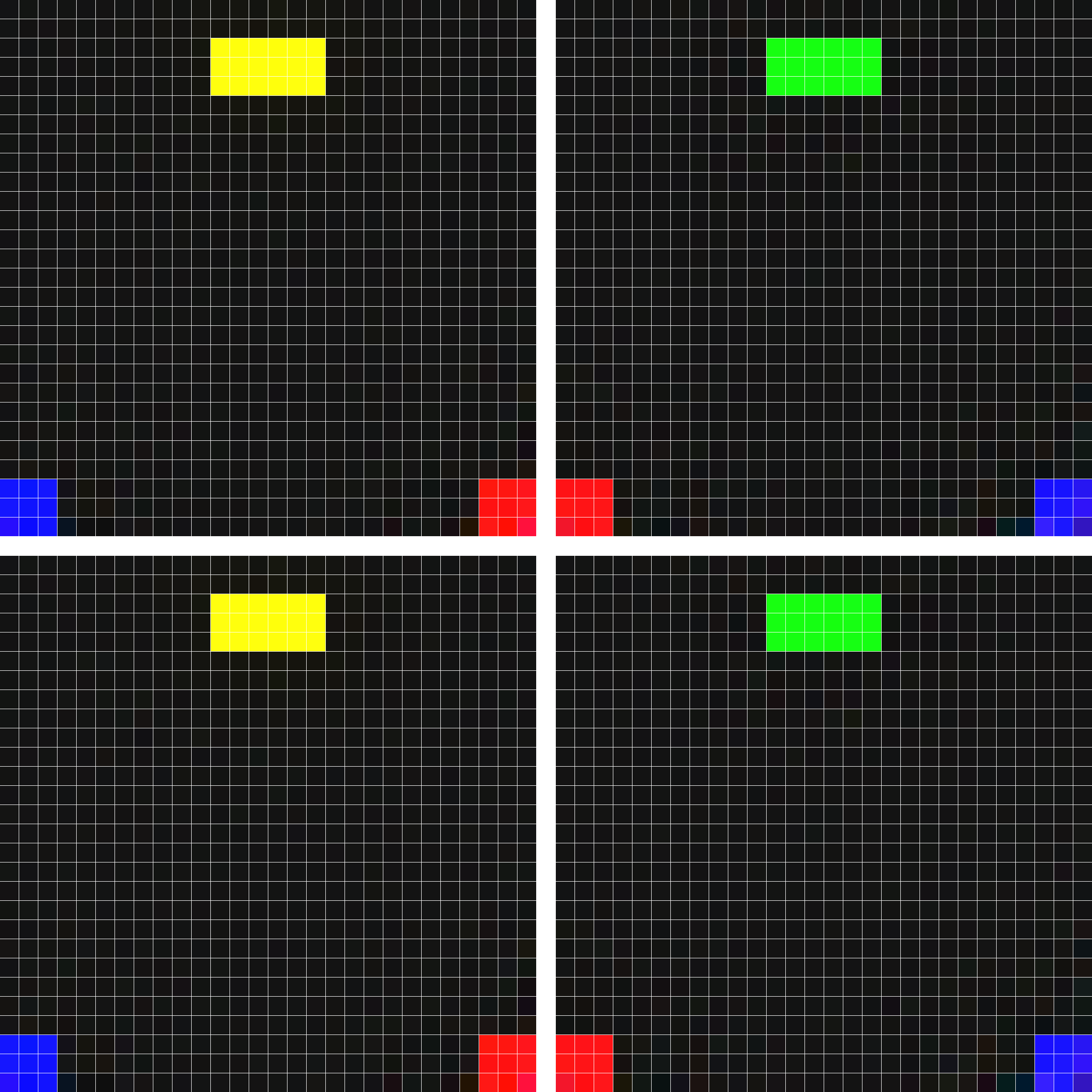}
        \caption{Epoch 90}
    \end{subfigure}%
    \begin{subfigure}{0.33\textwidth}
    \centering
        \includegraphics[width=0.95\linewidth]{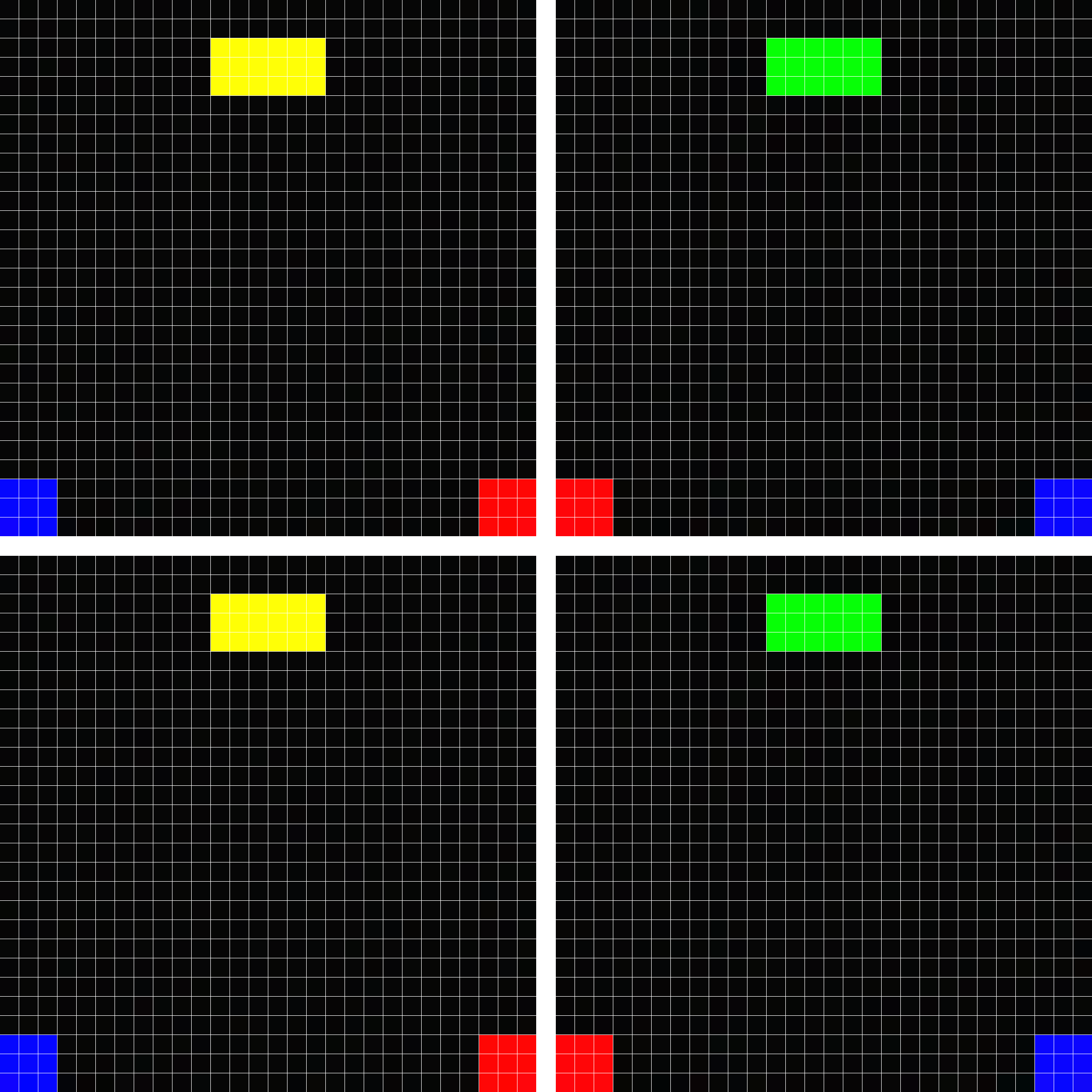}
        \caption{Epoch 120}
    \end{subfigure}%
    \caption{\textbf{Visualization of VAE training on BW-E} : After 30 epochs, VAE has learnt the general structure of the grid-world, but has yet to learn colors. After 60 epochs, it has learnt colors for marker pixels that were in most training examples, but has yet to learn goal state colors. After 90 epochs, it has learn colors for goal state, but is yet to learn colors at corners as they are present in very few samples. After 120 epochs, learning is complete.}
    \label{fig:vae_vis}
\end{figure}

To visualize the training of VAE, we use the same setup as for visualizing RBM training. We used 400 samples from BW-E and training was done in mini-batches of 128 samples. 4 test samples were randomly chosen and reconstruction from VAE was recorded after 30, 60, 90 and 120 epochs of training. Figure \ref{fig:vae_vis} shows the training progress of VAE on BW-E samples.
\section{Conclusions and Future Work}
We have presented a new method using a deep generative model to provide an exploration bonus for solving the multi-task reinforcement learning problem. Our modification to the VAE loss function allows it to learn from partial inputs and also the associations between different views of the same environment. Use of RBMs to learn a distribution over $\mathbf{z}$ allows us to sample actual MDPs instead of a mixture of MDPs. We introduced an intuitive exploration bonus and have shown improvements over existing baselines. \\
One drawback of Jacobian Bonus is that it doesn't use the reward structure of the MDPs. This bonus could yield sub-optimal policies in environments with multiple markers and associated rewards. We would like to incorporate the reward structure into the Jacobian Bonus to have some form of \emph{utility} interpretation.\\
Our deep generative model is scalable and we would like to explore learning in larger worlds and extend our method to work with Minecraft-like 3D environments.
\clearpage
\printbibliography
\end{document}